\documentclass[runningheads]{llncs}
\usepackage{graphicx}
\usepackage{tabularx}

\usepackage{dsfont}
\usepackage{booktabs}
\usepackage{enumerate}

\usepackage[centertags]{amsmath}
\usepackage{amssymb,amsfonts}

\everymath{\displaystyle}

\newcommand{\bx}{{\bf x}}
\newcommand{\by}{{\bf y}}

\newcommand{\bw}{{\bf w}}
\newcommand{\bm}{{\bf m}}
\newcommand{\ba}{{\bf a}}
\newcommand{\bb}{{\bf b}}

\usepackage{xcolor}

\definecolor{red}{RGB}{255,0,0}
\definecolor{black}{RGB}{0,0,0}
\definecolor{green}{RGB}{0, 255,0}
\definecolor{blue}{RGB}{0,0, 255}

\begin{document}
\title{Linear Dilation-Erosion Perceptron Trained Using a Convex-Concave Procedure\thanks{This work was supported in part by the S\~ao Paulo Research Foundation (FAPESP) under grant no 2019/02278-2, the National Council for Scientific and Technological Development (CNPq) under grant no 310118/2017-4, and the Coordena\c{c}\~ao  de Aperfei\c{c}oamento  de Pessoal de N\'ivel Superior - Brasil (CAPES) - Finance Code 001.}}
\author{Angelica Lourenço Oliveira\inst{1}\orcidID{0000-0002-8689-8522}\\ \and Marcos Eduardo Valle\inst{2}\orcidID{0000-0003-4026-5110}
}
\authorrunning{A. L. Oliveira \and M. E. Valle}

\institute{$^2$Campinas State University, Campinas SP 13083-970, Brazil \\
$^1$ \email{ra211686@ime.unicamp.br} \and \email{valle@ime.unicamp.br}}

\maketitle          % typeset the header of the contribution

\begin{abstract}
Mathematical morphology (MM) is a theory of non-linear operators used for the processing and analysis of images. Morphological neural networks (MNNs) are neural networks whose neurons compute morphological operators. Dilations and erosions are the elementary operators of MM. From an algebraic point of view, a dilation and an erosion are operators that commute respectively with the supremum and infimum operations. In this paper, we present the \textit{linear dilation-erosion perceptron} ($\ell$-DEP), which is given by applying linear transformations before computing a dilation and an erosion. The decision function of the $\ell$-DEP model is defined by adding a dilation and an erosion. Furthermore, training a $\ell$-DEP can be formulated as a convex-concave optimization problem. 
We compare the performance of the $\ell$-DEP model with other machine learning techniques using several classification problems. The computational experiments support the potential application of the proposed $\ell$-DEP model for binary classification tasks.

\noindent
\end{abstract}
\section{Introduction}\label{sec:intro}

Machine learning techniques play an important role in both pattern recognition and soft computing. Neural networks, which are inspired by the biological nervous system, are among the most efficient machine learning techniques used to solve pattern recognition tasks including, for example, computer vision and natural language processing \cite{Geron19HandsOn}. Due to page constraints, we will not provide an extensive review of the many interesting and effective neural network models which have been proposed in the last years but focus on the few models relevant to the development of the linear dilation-erosion perceptron addressed in this paper.

First of all, it is widely know that the perceptron, introduced by Rosenblatt in the late 1950s, can be used for binary classification tasks  \cite{rosenblatt58}. Precisely, the perceptron yields one class if the weighted sum of the inputs is greater than or equal to a threshold and returns the other class otherwise. Furthermore, Rosemblatt proposed a learning rule that converges whenever the samples of the two classes are linearly separable \cite{haykin_neural_2009}. The multi-layer perceptron (MLP) network, with at least one hidden layer and trained using backpropagation algorithms, overcomes the limitations of the Rosenblatt's perceptron for non-linearly separable classification tasks \cite{haykin_neural_2009}. The support vector classifiers (SVCs) developed by Vapnik and collaborators also overcomes the limitations of the perceptrons for binary classification tasks. Like the perceptron, an SVC also predicts the class of a sample by computing a weighted sum. However, training an SVC is performed by maximizing the margin of separation between the two classes \cite{vapnik_statistical_1998}. Furthermore, the classification performance of an SVC increases significantly by using the kernel trick. The kernel trick allows an SVC model to be effectively applied to non-linearly classification problems. For comparison purposes, in this paper, we consider both linear and radial-basis function SVC models as well as an MLP network.

In the late 1980s, Ritter and collaborators developed the so-called image algebra as an attempt to provide a unified framework for image processing and analysis techniques \cite{ritter90}. Using the image algebra framework, both linear and morphological operators are defined analogously. By replacing the usual linear dot-product in Rosemblatt's perceptron model with the corresponding lattice-based operation used to define morphological operators, Ritter and Sussner derived the morphological perceptron in the middle 1990s \cite{ritter96c,ritter97d}. The elementary operations from mathematical morphology are dilations and erosions \cite{soille99}. The maximum of additions yields a dilation while the minimum of sums yields an erosion. Hence, a dilation-based perceptron classifies a pattern by computing the maximum of its components plus the corresponding weights. Dually, the decision function of an erosion-based morphological perceptron is given by the minimum of the pattern's components plus the weights.

Such as the MLP, a multi-layer morphological perceptron (MLMP) with at least one hidden layer with both dilation-based and erosion-based neurons can theoretically solve any binary classification tasks \cite{ritter03}. From a geometric point of view, an MLMP discriminates two classes by enclosing patterns in hyper-boxes \cite{ritter03,Sussner1998MorphologicalLearning}. The non-differentiability of the lattice-based operations, however, makes training morphological neural networks difficult for gradient-based algorithms. Thus, some researchers developed training algorithms in which the network structure grows by adding hyper-boxes to fit the training data \cite{ritter03,Sussner1998MorphologicalLearning,sussner_morphological_2011}. On the downside, growing algorithms usually overfit the training data. 

Recently, Charisopoulos and Maragos formulated the training of a morphological perceptron as a concave-convex optimization problem which, in some sense, resembles the training of a linear SVC \cite{charisopoulos_morphological_2017}. Despite the encouraging performance in simple classification tasks, both dilation-based and erosion-based morphological perceptrons implicitly assume a relationship between the input features and their classes \cite{valle_reduced_2020}. Using concepts from vector-valued mathematical morphology, one of us proposed the reduced dilation-erosion perceptron (r-DEP), which overcomes the limitations of the morphological perceptron \cite{valle_reduced_2020}. In a few words, the r-DEP first transforms the input into a feature vector. The feature vector is then classified using a linear combination of a dilation-based and an erosion-based perceptron. 

In this paper, a linear transformation is used to map the input to the feature vector. The resulting model is referred to as a \textit{linear dilation-erosion perceptron} ($\ell$-DEP). We point out that a $\ell$-DEP model can also theoretically solve any binary classification task. Furthermore, inspired by the works of Charisopoulos and Maragos, we formulate the training of a $\ell$-DEP classifier as a convex-concave optimization problem. 

The paper is organized as follows. The next section provides the mathematical background necessary to understand the models presented in this paper. Section \ref{sec:rdep} briefly reviews the morphological perceptron as well as the $r$-DEP model. The $\ell$-DEP model and its training algorithm are introduced in Section \ref{sec:l-DEP}.  Computational experiments comparing the performance of the $\ell$-DEP model with other classical models from the literature are given in Section \ref{comp_exp}. The paper finishes with some concluding remarks in Section \ref{sec:concl}.

    \section{Basic Concepts on Mathematical Morphology}\label{sec:bas_conc}
    
    Mathematical morphology is a non-linear theory widely used for image processing and analysis \cite{heijmans95,soille99}. From the theoretical point of view, mathematical morphology is very well defined on complete lattices.
    A partially ordered set ($\mathbb{L},\preceq$) is a { complete lattice} if any subset of $\mathbb{L}$ has both a supremum and an infimum, which are denoted respectively by $ \sup X$ and $\inf X$. 
    
    The elementary operators from mathematical morphology commute with the lattice-based operations. Formally, given complete lattices $\mathbb{L}$ and $\mathbb{M}$, a dilation $\delta:\mathbb{L} \to \mathbb{M}$ and an erosion $\varepsilon:\mathbb{L} \to \mathbb{M}$ satisfy respectively the following identities for all $X \in \mathbb{L}$ \cite{heijmans90}: 
    \begin{equation}\label{Eq:erodil}
    \delta\left(\sup X\right)= \sup_{\bx \in X} \{\delta(\bx)\} \quad \text{and} \quad \varepsilon\left(\inf X\right)= \inf_{\bx \in X} \{\varepsilon(\bx)\}.
    \end{equation}
    
    \begin{example}
    Let $\bar{\mathbb{R}} = \mathbb{R} \cup \{-\infty,+\infty\}$ denote the extended real-numbers. The cartesian product $\bar{\mathbb{R}}^n$ is a complete lattice with the partial ordering given by $\bx \preceq \by  \Leftrightarrow x_i \leq y_i, \forall i =1:n.$    Given $\ba,\bb \in \mathbb{R}^n$, the operators $\delta_{\ba},\varepsilon_{\bb}:\bar{\mathbb{R}}^n \to \bar{\mathbb{R}}$ given by
    \begin{equation} 
    \delta_{\ba}(\bx)=\max_{j=1:n}\{a_j + x_j\} \quad \mbox{and} \quad  \varepsilon_{\bb}(\bx)=\min_{j=1:n} \{b_j + x_j\}, \end{equation}  
    for all $\bx\in\mathbb{\bar{R}}^n$, are respectively a dilation and an erosion \cite{sussner_morphological_2011}.
    \end{example}
    
    The lattice-based elementary operations of mathematical morphology can be extended to more abstract sets using the concept of reduced orderings \cite{goutsias_morphological_1995}. Let us briefly review the concepts of reduced dilation and reduced erosion as defined recently in \cite{valle_reduced_2020}.
    Let $\mathbb{V}$ and $\mathbb{W}$ be an arbitrary sets, not necessarily complete lattices. Also, let $\rho:\mathbb{V} \to \mathbb{L}$ and $\sigma:\mathbb{W} \to \mathbb{M}$ be surjective mappings, where $\mathbb{L}$ and $\mathbb{M}$ are both complete lattices. An operator $\delta^r:\mathbb{V} \to \mathbb{W}$ is a reduced dilation and an operator $\varepsilon^r:\mathbb{V} \to \mathbb{W}$ is a reduced erosion if there exist a dilation $\delta:\mathbb{L} \to \mathbb{M}$ and an erosion $\varepsilon: \mathbb{L} \to \mathbb{M}$ such that
    \begin{equation} \sigma \left(\delta^r(\bx) \right) = \delta \left( \rho(\bx) \right) \quad \mbox{and} \quad 
    \sigma \left(\varepsilon^r(\bx) \right) = \varepsilon \left( \rho(\bx) \right), \quad  \forall \bx \in \mathbb{V}. \end{equation}
    Reduced morphological operators, such as the reduced dilation and the reduced erosion defined above, have been effectively used for processing vector-valued imagens such as color and hyperspectral images \cite{velasco-forero_supervised_2011,velasco-forero14}.
    
    \section{Reduced Dilation-Erosion Perceptron}\label{sec:rdep}
    
    Morphological neural networks (MNNs) are neural networks whose  induced local field of the neurons are given by an operation from mathematical morphology \cite{sussner_morphological_2011}. The morphological perceptrons, introduced by Ritter and Sussner in the middle 1990s for binary classification tasks, are one of the earliest MNNs \cite{ritter96c}. In a few words, the morphological perceptrons are obtained by replacing the usual affine transformation $A(\bx) = \langle \mathbf{w}, \bx \rangle + b$, for all $\bx \in \mathbb{R}^n$, by an elementary morphological operator in Rosenblatt's perceptron given by $y = f\big(A(\bx))\equiv fA(\bx)$, where $f$ denotes a hard limiter activation function \cite{haykin99}. Formally, the two morphological perceptrons are given by the equations $y = f\big(\delta_{\ba}(\bx)\big) \equiv f \delta_{\ba}(\bx)$ and $y = f\big(\varepsilon_{\bb}(\bx)\big) \equiv f\varepsilon_{\bb}(\bx)$, for all $\bx \in \bar{\mathbb{R}}^n$. Specifically, the dilation-based perceptron and the erosion-based perceptron are given respectively by
	\begin{equation}\label{Eq:perc_ero_dil}
	y=f\left(\max_{j=1:n}\{a_j + x_j\}\right)\quad \text{ or } \quad	y=f\left(\min_{j=1:n}\{b_j + x_j\}\right), \quad \bx \in \bar{\mathbb{R}}^n.
	\end{equation}  
	For simplicity, in this paper we shall consider the signal function as the hard limiter activation function of the morphological perceptrons, i.e., $f(x) = +1$ if $x \geq 0$ and $f(x)=0$, otherwise. 
	Note that a morphological perceptron model is given by either the composition $f\delta_{\ba} $ or the composition $f \varepsilon_{\bb}$, where $\delta_{\ba} : \mathbb{\bar{R}}^n \rightarrow \mathbb{\bar{R}}$ and $\varepsilon_{\bb} : \mathbb{\bar{R}}^n \rightarrow \mathbb{\bar{R}}$ denote respectively the dilation and the erosion given  \eqref{Eq:erodil}, for $\ba,\bb\in \mathbb{R}^n$. Because of the maximum operation, a dilation-based perceptron given by $y = f\delta_{\ba}$ favors the positive class whose label is $+1$ \cite{valle_reduced_2020}. Dually, an erosion-based perceptron defined by $y = f \varepsilon_{\bb}$ favors the negative class, i.e., the class label $-1$, because of the minimum operation.
    
    In \cite{de_a_araujo_class_2011}, Araujo proposed an hybrid MNN called dilation-erosion perceptron (DEP). For binary classificaiton tasks, a DEP computes the convex combination of a dilation and an erosion followed by a hard-limiter function. Intuitively, the DEP model allows for a balance between the two classes. In mathematical terms, given $\beta \in [0,1]$, a DEP model is defined by $y=f(\tau(\bx)) \equiv f \tau(\bx)$, where
    \begin{align}\label{Eq:dep}
        \tau(\bx) = \beta\delta_{\ba}+(1-\beta) \varepsilon_{\bb},\quad \forall \bx \in \mathbb{R}^n,
    \end{align}
    is the decision function of the DEP model.
 
    Despite the successful application of the DEP model for times series prediction \cite{de_a_araujo_class_2011}, the DEP classifiers have a serious drawback: as a lattice-based model, the DEP classifier presupposes a partial ordering on the input space as well as on the set of classes. 
    To overcome this problem, Valle recently proposed the so-called {\it reduced dilation-erosion perceptron} ($r$-DEP) using concepts from vector-valued mathematical morphology \cite{valle_reduced_2020}. An $r$-DEP classifier is defined as follows: 
    Let $\mathbb{V}$ be the input space and let $\mathbb{C}=\{c_1,c_2\}$ be the set of class labels of a binary classification task. The input space $\mathbb{V}$ is usually a subset of $\mathbb{R}^n$, but we may consider more abstract input spaces. Also, consider the complete lattice $\mathbb{L} = \mathbb{\bar{R}}^r$, equipped with the usual component-wise ordering, and let us denote $\mathbb{M} = \{-1,+1\}$. Given a one-to-one correspondence ${\sigma: \mathbb{C} \rightarrow \mathbb{M}}$ and a surjective mapping $\rho: \mathbb{V} \to \bar{\mathbb{R}}^r$,
    an $r$-DEP model is defined by the equation $y=\sigma^{-1}\big(f\big(\tau^r(\bx)\big)\big) \equiv \sigma^{-1} f \tau^r (\bx)$, where $\tau^r:\mathbb{V} \to \mathbb{R}$ is the decision function given by the following convex combination for $\beta \in [0,1]$:
    \begin{equation} \label{eq:tau-r}
        \tau^r(\bx) = \beta \delta_{\ba}\big(\rho(\bx)\big)+(1-\beta) \varepsilon_{\bb}\big(\rho(\bx)\big), \quad \forall \bx \in \mathbb{V}.
    \end{equation}
    
    Finally, let us briefly comment on the training of an {$r$-DEP} classifier. Consider a training set $\mathcal{T}=\{(\bx_i,d_i):i=1,\ldots,m\} \subseteq \mathbb{V} \times \mathbb{C}$. From equation \eqref{eq:tau-r}, training an $r$-DEP is done simply training a DEP classifier using the transformed training data $\mathcal{T}^r=\{(\rho(\bx_i),\sigma(d_i)):i=1,\ldots,m\} \subseteq \mathbb{L}\times\mathbb{M}$. At this point, we would like to recall that Charisopoulos and Maragos   formulated the training of the morphological perceptrons as well as the hybrid DEP classifier as the solution of a convex-concave programming (CCP) problem \cite{charisopoulos_morphological_2017}. Training $r$-DEP classifier using convex-concave procedures yielded encouraging results on classification tasks \cite{valle_reduced_2020}. 
    
	\section{Linear Dilation-Erosion Perceptron}\label{sec:l-DEP}
	
	The main challenge in defining a successful {$r$-DEP} classifier is how to determine the surjective mapping $\rho:\mathbb{V} \to \mathbb{R}^r$. In this paper, we propose linear transformations as the mapping $\rho$, resulting in the so-called {\it linear dilation-erosion perceptron} ($\ell$-DEP) classifier. Precisely, a $\ell$-DEP model is defined as follows. Let the input space be the $n$-dimensional vector space $\mathbb{V} = \mathbb{R}^n$ and let $\mathbb{C}=\{c_1,c_2\}$ be the set of class labels of a binary classification task. 
    Also, let $\mathbb{L}_1=\mathbb{\bar{R}}^{r_1}$ and $\mathbb{L}_2=\mathbb{\bar{R}}^{r_2}$ be complete lattices with the usual component-wise partial ordering. Given linear mappings $\rho_1 : \mathbb{V} \to \mathbb{L}_1$ and $\rho_2 : \mathbb{V} \to  \mathbb{L}_2$, the application of linear mappings before the evaluation of an elementary morphological mapping yields the decision function $\tau^l:\mathbb{V} \to \mathbb{R}$ given by  
    \begin{equation} \label{eq:tau-l}
        \tau^l(\bx) = \beta \delta_{\ba}\big(\rho_1(\bx)\big) + (1-\beta) \varepsilon_{\bb}\big(\rho_2(\bx)\big), 
        \quad \bx \in \mathbb{V}.\nonumber
     \end{equation}
    Finally, such as the previous model, the $\ell$-DEP model is defined by the equation \begin{equation} y=\sigma^{-1}\big(f\big(\tau^l(\bx)\big)\big) \equiv \sigma^{-1} f \tau^l (\bx), \end{equation} 
    where $f$ is the sign function and $\sigma:\mathbb{C}\to \{-1,+1\}$ is a one-to-one mapping.
    
    In contrast to the decision function $\tau^r$ of the $r$-DEP model given by \eqref{eq:tau-r}, the decision function $\tau^l$ of the $\ell$-DEP is defined using two linear mappings $\rho_1$ and $\rho_2$. As a consequence, $\tau^l$ is a continuous piece-wise linear function which is able to approximate any continuous-valued function from a compact on $\mathbb{R}^n$ to $\mathbb{R}$ \cite{goodfellow_maxout_2013,stone_generalized_1948}. Let us formalize this remark. The linear mappings $\rho_1$ and $\rho_2$ satisfies $\rho_1(\bx) = R_1\bx$ and $\rho_2(\bx) = R_2 \bx$ for real-valued matrices $R_1 \in \mathbb{R}^{r_1 \times n}$ and $R_2 \in \mathbb{R}^{r_2 \times n}$. Defining $W = \beta R_1 \in \mathbb{R}^{r_1 \times n}$, $M = (\beta-1)R_2 \in \mathbb{R}^{r_2 \times n}$, $\mathbf{c} = \beta \ba$, and $\mathbf{d} = (\beta-1) \bb$, the decision function of the $\ell$-DEP classifier can be alternatively written as
    \begin{align}
        \tau^l(\bx) &= \beta \delta_{\ba}(R_1 \bx) + (1-\beta) \varepsilon_{\bb}(R_2 \bx) \nonumber \\
        &= \beta \max\{R_1 \bx + \ba\} - (\beta-1) \min\{R_2 \bx + \bb\} \nonumber \\
        &= \max\{\beta R_1 \bx + \beta \ba\} -  \max\{(\beta-1)R_2 \bx + (\beta-1)\bb\} \nonumber \\
        &= \max\{W \bx + \mathbf{c}\} -  \max\{M \bx + \mathbf{d}\} \nonumber \\ 
        &= \max_{i=1:r_1}\{\bw_i^T \bx + c_i\} -  \max_{j=1:r_2}\{\bm_j^T \bx + d_j\}, \label{eq:tau-DC}
    \end{align}
    where $\bw_i^T$ and $\bm_i^T$ are rows of $W$ and $M$, respectively. From the last identity, we can identify $\tau^l$ with a piece-wise linear function \cite{wang_general_2004}. Moreover, from Theorem 4.3 in \cite{goodfellow_maxout_2013},  $\tau^l$ is an universal approximator. Thus, a $\ell$-DEP classifier can theoretically solve any binary classification problem. Let us now address the training of a $\ell$-DEP classifier. 

	Given a training dataset $T=\{(\bx_k,d_k):k=1,\ldots,m \} \subseteq \mathbb{V} \times \mathbb{C}$, define the sets $C^+ = \{\bx_k: \sigma(d_k) = +1\}$ and $C^- = \{\bx_k: \sigma(d_k) = -1\}$ of input samples. Inspired by the linear SVC and the concave-convex procedure developed by Charisopoulos and Maragos \cite{charisopoulos_morphological_2017,haykin_neural_2009}, the parameters $W$, $\mathbf{c}$, $M$ and $\mathbf{d}$ of a $\ell$-DEP classifier can be determined by minimizing the hinge loss function subject to the constraints
	\begin{equation}
	    \tau^l(\bx_k) \leq -1+\xi_k, \text{ if } \bx_k \in C^-, \quad \text{and} \quad \tau^l(\bx_k) \geq 1-\xi_k,  \text{ if } \bx_k \in C^+.
	\end{equation}
   The terms $-1$ and $+1$ in the left-hand side of the inequalities imposes a margin of separation between the two classes. The slack variables $\xi_k$, for $ k = 1, \ldots, m $, allows some classification errors in the training set. Note that, regardless of its class, the input sample $ \bx_k$ is well classified if $\xi_k \leq 0$. The hinge loss minimizes the positive values of $\xi_k$, for $\xi=1,\ldots,m$.

    Concluding, from \eqref{eq:tau-DC}, a $\ell$-DEP classifier is trained by solving the following disciplined convex-concave programming (DCCP) problem \cite{yuille_concave-convex_2003}:
    \begin{equation} \label{eq:DCCP}
        \begin{cases}
        \mathop{\mbox{minimize}}_{W,\mathbf{c},M,\mathbf{d},\boldsymbol{\xi}} & \sum_{k=1}^m \max(\xi_k,0), \\ 
          \mbox{s.t.} & \max_{i=1:r_1}(\bw_i^T \bx_k + c_i) + 1\,\, \leq \max_{j=1:r_2}(\bm_j^T \bx_k + d_j) + \xi_k, \forall \bx_k \in C^- \\
          & \max_{i=1:r_1}(\bw_i^T \bx_k + c_i) + \xi_k \geq \max_{j=1:r_2}(\bm_j^T \bx_k + d_j) + 1,  \,\,\forall \bx_k \in C^+
        \end{cases}
    \end{equation}
    We would like to point out that we solved the disciplined convex-concave programming problem \eqref{eq:DCCP} using the algorithm proposed by Shen et. al. \cite{shen_disciplined_2016}. Precisely, in our computational implementations, we solved \eqref{eq:DCCP} using the \texttt{CVXPY} package \cite{diamond_cvxpy_2016}, which has an extension for solving DCCP problems, combined with the \texttt{MOSEK} solver \cite{aps_mosek_2020}.

    \section{Computational Experiments}\label{comp_exp}
    
    Let us briefly evaluate the performance of the proposed $\ell$-DEP classifier on several datasets from the OpenML repository available at \url{https://www.openml.org/}. For simplicity, we fixed the parameters $r_1 = r_2 = 10$ of the $\ell$-DEP classifier for all datasets. Let us also compare the proposed classifier with a multi-layer perceptron (MLP), the maxout network, and the linear and the RBF SVCs \cite{goodfellow16book,goodfellow_maxout_2013,haykin99}. We used \texttt{python’s scikit-learn (\texttt{sklearn})} implementations of the MLP and SVC classifiers with their default parameters  \cite{pedregosa_scikit-learn_2011}. The maxout network has been implemented using \texttt{tensorflow} with extra functionalities, that is, the maxout layer of \texttt{tensorflow-addons}. In our experiments, we used $10$ locally affine regions in the maxout network for all datasets \cite{goodfellow_maxout_2013}.
    
     We would like to point out that we handled missing data using sklearn's \texttt{SimpleImputer()} command. Furthermore, we partitioned the data set into training and test sets using the sklearn's \texttt{StratifiedKFold()} command with $k=5$. 
     Finally, since some datasets are not balanced, we used the F1-score to measure the performance of the classifiers.
    
\begin{figure}\label{fig:box_scores_times}
    \centering
        \includegraphics[width=.49\columnwidth]{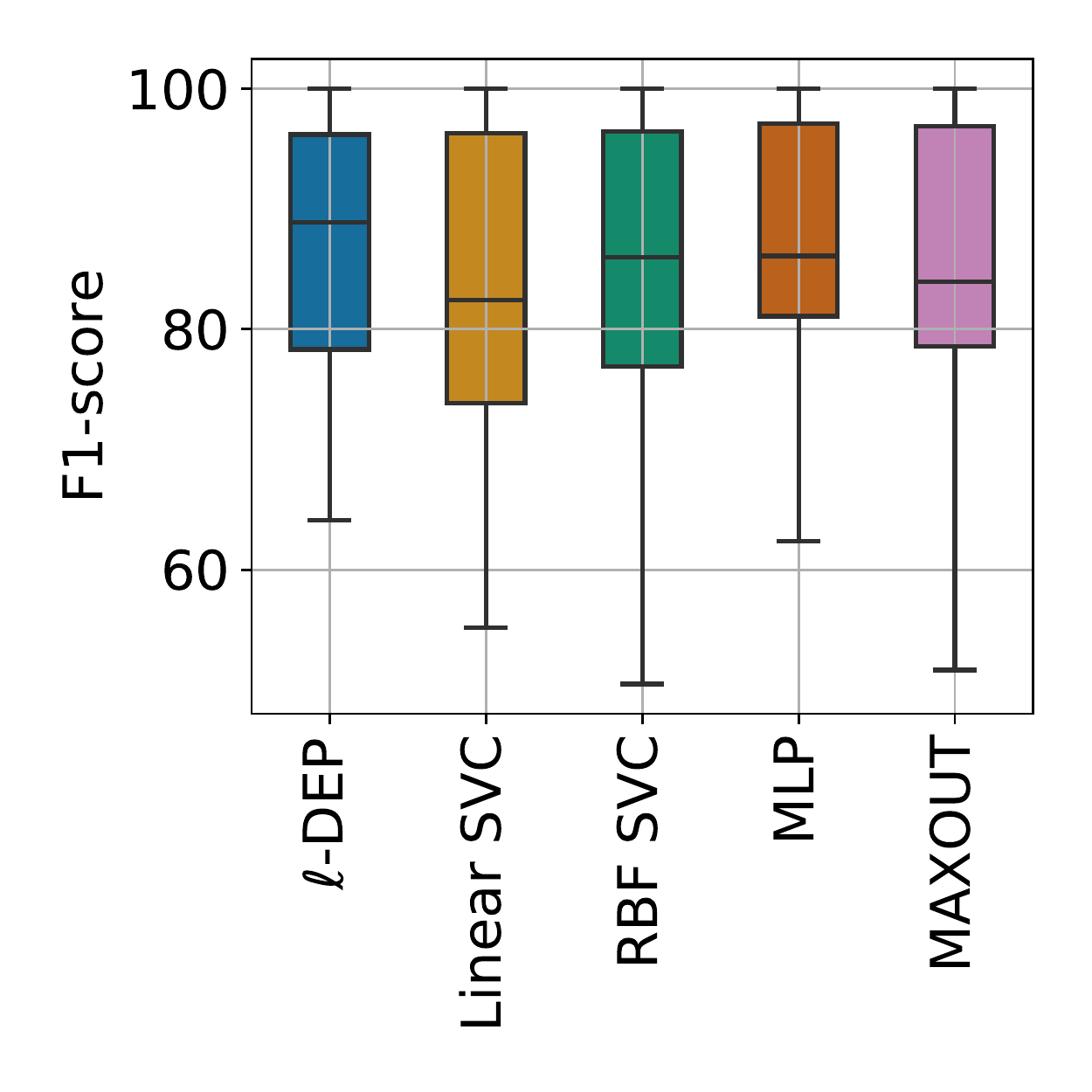}
        \includegraphics[width=.49\columnwidth]{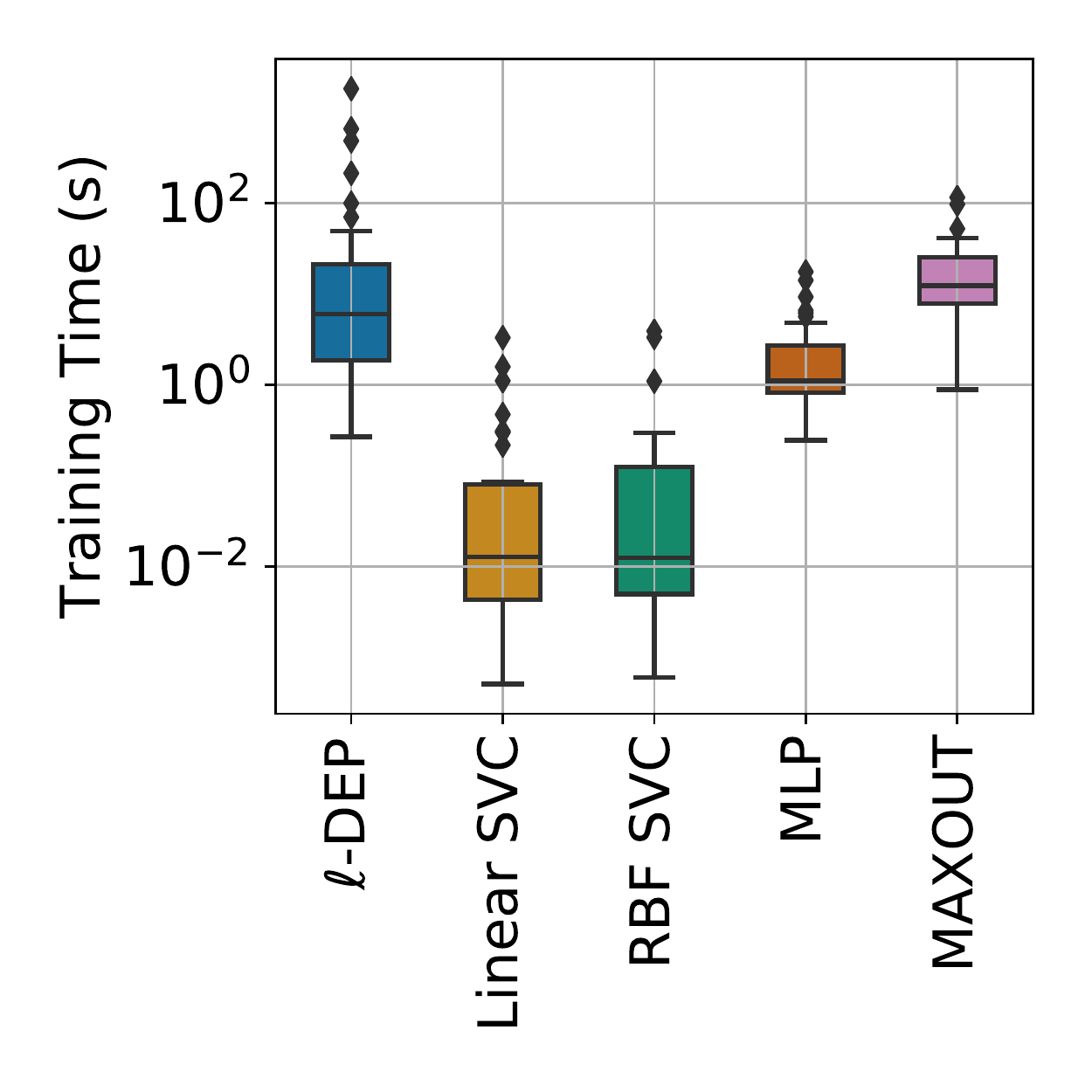}
        \caption{Boxplot of the F1-score and time required to train the binary classifiers}
    \end{figure}

    Table \ref{tab:accu} contains the mean and the standard deviation of the F1-score obtained from the five classifiers using stratified $5$-fold cross-validation. The boxplot shown in Figure \ref{fig:box_scores_times} summarizes the scores depicted on this table as well as the execution time for training the clasifiers.
    Although the MPL yielded the largest average performance, the $\ell$-DEP model produced the largest median performance. Furthermore, it is clear from the boxplot on the left of Figure \ref{fig:box_scores_times} that the $\ell$-DEP is comparable to the other classifiers. In particular, $\ell$-DEP achieved slightly higher performance than the maxout network. 
    As to the training execution time, the linear and the RBF SVCs are the fasted models. Despite its longer, the $\ell$-DEP is not quite different from the MLP and maxout models in training time.

    \begin{table}[t]
    \begin{center}
    \caption{Average and standard deviation of the F1-Score.}
    \label{tab:accu}
    \resizebox{\textwidth}{!}{
    {\scriptsize
    \begin{tabular}{l||c|c|c|c|c}
        \toprule
        {} &                    $\ell$-DEP &              Linear SVC &                 RBF SVC &                       MLP &                  MAXOUT \\
        \midrule
        Accute Inflammations    &  \textbf{100 \,$\pm$ 0.0} &  \textbf{100 \,$\pm$ 0.0} &  \textbf{100 \,$\pm$ 0.0} &    \textbf{100 \,$\pm$ 0.0} &  \textbf{100 \,$\pm$ 0.0} \\
        Australian              &            83.3 $\pm$ 1.9 &            85.2 $\pm$ 3.6 &   \textbf{86.1 $\pm$ 2.9} &     \textbf{86.1 $\pm$ 1.9} &            85.5 $\pm$ 3.1 \\
        Banana                  &            89.2 $\pm$ 3.1 &            55.2 $\pm$ 0.0 &   \textbf{90.4 $\pm$ 1.1} &              90.1 $\pm$ 1.4 &           84.0 $\pm$ 10.4 \\
        Banknote                &  \textbf{100 \,$\pm$ 0.0} &            98.5 $\pm$ 0.5 &  \textbf{100 \,$\pm$ 0.0} &    \textbf{100 \,$\pm$ 0.0} &            99.5 $\pm$ 1.0 \\
        Blood Transfusion       &            77.7 $\pm$ 2.4 &            76.1 $\pm$ 0.5 &            76.6 $\pm$ 1.1 &     \textbf{79.8 $\pm$ 1.5} &            79.5 $\pm$ 2.0 \\
        Breast Cancer Wisconsin &            96.7 $\pm$ 1.1 &            97.2 $\pm$ 0.7 &   \textbf{97.7 $\pm$ 0.5} &              97.5 $\pm$ 1.4 &            97.2 $\pm$ 0.7 \\
        Chess                   &   \textbf{99.3 $\pm$ 0.4} &            96.8 $\pm$ 0.6 &            98.3 $\pm$ 0.7 &              99.2 $\pm$ 0.4 &            99.1 $\pm$ 0.1 \\
        Colic                   &            81.5 $\pm$ 4.0 &            81.3 $\pm$ 6.6 &   \textbf{85.1 $\pm$ 3.7} &              84.8 $\pm$ 3.2 &            81.6 $\pm$ 5.9 \\
        Credit Approval         &            80.4 $\pm$ 3.3 &            85.5 $\pm$ 2.8 &   \textbf{85.7 $\pm$ 3.0} &              84.9 $\pm$ 0.9 &            83.9 $\pm$ 2.8 \\
        Credit-g                &            68.4 $\pm$ 5.0 &            75.8 $\pm$ 3.0 &   \textbf{76.5 $\pm$ 4.4} &              73.8 $\pm$ 3.7 &            68.5 $\pm$ 4.6 \\
        Cylinder Bands          &            76.7 $\pm$ 5.9 &            69.4 $\pm$ 2.4 &            77.2 $\pm$ 2.8 &     \textbf{77.4 $\pm$ 5.8} &            72.8 $\pm$ 4.7 \\
        Diabetes                &            70.8 $\pm$ 3.6 &            76.4 $\pm$ 3.4 &            75.2 $\pm$ 4.7 &     \textbf{76.7 $\pm$ 3.7} &            75.1 $\pm$ 3.3 \\
        EEG-Eye-State           &   \textbf{89.3 $\pm$ 1.5} &            61.7 $\pm$ 1.9 &            68.1 $\pm$ 5.0 &              85.6 $\pm$ 3.1 &            74.1 $\pm$ 7.5 \\
        Haberman                &            70.3 $\pm$ 3.8 &            73.2 $\pm$ 0.8 &            73.5 $\pm$ 2.6 &     \textbf{74.5 $\pm$ 2.4} &            74.2 $\pm$ 1.8 \\
        Hill-Valley             &   \textbf{93.6 $\pm$ 1.3} &            60.6 $\pm$ 1.9 &            50.5 $\pm$ 1.2 &              62.4 $\pm$ 2.7 &            51.7 $\pm$ 2.0 \\
        Ilpd                    &            64.1 $\pm$ 4.8 &            71.4 $\pm$ 0.4 &            71.2 $\pm$ 0.3 &     \textbf{73.1 $\pm$ 3.0} &            70.2 $\pm$ 3.1 \\
        Internet Advertisements &            93.7 $\pm$ 1.8 &            96.0 $\pm$ 0.7 &            96.6 $\pm$ 0.3 &     \textbf{97.2 $\pm$ 0.4} &            97.0 $\pm$ 0.6 \\
        Ionosphere              &            88.9 $\pm$ 4.5 &            88.6 $\pm$ 2.7 &   \textbf{94.3 $\pm$ 2.3} &              92.6 $\pm$ 3.1 &            91.7 $\pm$ 4.5 \\
        MOFN-3-7-10             &  \textbf{100 \,$\pm$ 0.0} &  \textbf{100 \,$\pm$ 0.0} &  \textbf{100 \,$\pm$ 0.0} &    \textbf{100 \,$\pm$ 0.0} &  \textbf{100 \,$\pm$ 0.0} \\
        Monks-2                 &   \textbf{88.0 $\pm$ 3.7} &            65.7 $\pm$ 0.2 &            72.7 $\pm$ 2.3 &              82.4 $\pm$ 3.1 &            83.5 $\pm$ 3.2 \\
        Mushroom                &  \textbf{100 \,$\pm$ 0.0} &            97.8 $\pm$ 0.9 &  \textbf{100 \,$\pm$ 0.0} &    \textbf{100 \,$\pm$ 0.0} &  \textbf{100 \,$\pm$ 0.0} \\
        Phoneme                 &            84.9 $\pm$ 1.2 &            77.3 $\pm$ 1.6 &            84.5 $\pm$ 1.5 &     \textbf{85.6 $\pm$ 1.4} &            81.5 $\pm$ 3.5 \\
        Pishing Websites        &            95.7 $\pm$ 0.6 &            90.5 $\pm$ 0.7 &            95.0 $\pm$ 0.9 &     \textbf{96.9 $\pm$ 0.5} &            95.6 $\pm$ 0.6 \\
        Sick                    &            97.0 $\pm$ 0.7 &            96.6 $\pm$ 1.0 &            96.3 $\pm$ 0.4 &     \textbf{97.1 $\pm$ 0.7} &            96.8 $\pm$ 1.0 \\
        Sonar                   &            77.4 $\pm$ 3.5 &            74.5 $\pm$ 3.9 &   \textbf{84.2 $\pm$ 4.2} &     \textbf{84.2 $\pm$ 3.5} &            80.8 $\pm$ 4.4 \\
        Spambase                &            92.8 $\pm$ 1.5 &            92.9 $\pm$ 0.5 &            93.3 $\pm$ 0.5 &     \textbf{94.5 $\pm$ 0.9} &            93.5 $\pm$ 0.6 \\
        Steel Plates Fault      &  \textbf{100 \,$\pm$ 0.0} &  \textbf{100 \,$\pm$ 0.0} &            99.6 $\pm$ 0.5 &              99.9 $\pm$ 0.1 &            99.9 $\pm$ 0.1 \\
        Thoracic Surgery        &            76.8 $\pm$ 3.1 &   \textbf{85.1 $\pm$ 0.0} &   \textbf{85.1 $\pm$ 0.0} &              83.6 $\pm$ 1.8 &            81.9 $\pm$ 2.7 \\
        Tic-Tac-Toe             &            92.6 $\pm$ 5.0 &            65.3 $\pm$ 0.2 &            88.8 $\pm$ 2.1 &              86.3 $\pm$ 3.2 &   \textbf{95.3 $\pm$ 4.5} \\
        Titanic                 &   \textbf{79.0 $\pm$ 2.3} &            77.6 $\pm$ 2.0 &            77.8 $\pm$ 1.9 &              77.7 $\pm$ 1.9 &            77.6 $\pm$ 1.7 \\
        \bottomrule
        MEAN $\pm$ STD                 &           86.9 $\pm\,$ 10.6 &           82.4 $\pm$ 13.3 &           86.0 $\pm$ 12.0 &  \textbf{87.5 $\pm$ 10.0} &           85.7 $\pm$ 12.0 
        \\ 
        MEDIAN $\pm$ MAD & \textbf{88.9 $\pm$ 9.0} & 82.4 $\pm$ 11.3 & 86.1 $\pm$ 9.4& 86.1 $\pm$ 8.3 & 84.0 $\pm$ 9.8  
    \end{tabular}
    }}
    \end{center}
    \end{table} 
    
\section{Concluding Remarks}\label{sec:concl}

    In this paper, we introduced the linear dilation-erosion perceptron ($\ell$-DEP) classifier which is given by a convex combination of the composition of linear transformations and the two elementary morphological operation. Specifically, given a one-to-one mapping $\sigma$ from the set of class labels $\mathbb{C}$ to $\{+1,-1\}$, a $\ell$-DEP classifier is defined by means of the equation $y=\sigma^{-1} f \tau^l(\bx)$, where the decision function $\tau^l:\mathbb{R}^n \to \mathbb{R}$ is given by \eqref{eq:tau-l}. Alternatively, $\tau^l$ can be expressed by means of \eqref{eq:tau-DC} for matrices $W \in \mathbb{R}^{r_1 \times n}$ and $M \in \mathbb{R}^{r_2 \times n}$ and vectors $\mathbf{c} \in \mathbb{R}^{r_1}$ and $\mathbf{d} \in \mathbb{R}^{r_2}$. Except for the one-to-one mapping $\sigma:\mathbb{C} \to \{-1,+1\}$, which can be defined arbitrarily, the other parameters $W$, $M$, $\mathbf{c}$, and $\mathbf{d}$ are self-adjusted in the training of the $\ell$-DEP model. Moreover, the classifier is trained by solving the disciplined convex-concave programming problem given by \eqref{eq:DCCP}, which has been inspired by the works of Charisopoulos and Maragos \cite{charisopoulos_morphological_2017}. From a theoretical point of view, the decision function $\tau^l$ of the $\ell$-DEP is a continuous piece-wise linear function \cite{wang_general_2004}. As a consequence, a $\ell$-DEP model can in principle solve any binary classification task \cite{goodfellow_maxout_2013}. Computational experiments with 30 binary classification problems revealed comparable performance of the proposed $\ell$-DEP model with other classifiers from the literature, namely, the linear and RFB SVCs, MLP, and the maxout network. 

    In the future, we plan to investigate further in detail the concave-convex programming problem used to train a $\ell$-DEP classifier. In particular, we intend to include regularization terms in the objective function to improve the generalization capability of the $\ell$-DEP classifier.

\bibliographystyle{splncs04}
\bibliography{references,references2}

\end{document}